%% file: main.tex
\documentclass[10pt,twocolumn,letterpaper]{article}

\usepackage{cvpr}              

\usepackage{graphicx}
\usepackage{amsmath}
\usepackage{amssymb}
\usepackage{booktabs}
\usepackage{relsize}
\usepackage{multirow}
\usepackage{pifont}
\usepackage{setspace}

\usepackage[pagebackref,breaklinks,colorlinks]{hyperref}

\usepackage[capitalize, noabbrev]{cleveref}
\Crefname{section}{Section}{Sections}
\Crefname{table}{Table}{Tables}

\def\cvprPaperID{2867} 
\def\confName{CVPR}
\def\confYear{2023}

\newcommand{\vectorname}[1]{{\mathrm{\mathbf{#1}}}}
\newcommand{\myparagraph}[1]{\vspace{2.35pt}\noindent{\bf #1:}}
\newcommand{\keypoint}[1]{\vspace{0.1cm}\noindent\textbf{#1}\quad}

\begin{document}

\title{Data-Free Sketch-Based Image Retrieval}



\author{Abhra Chaudhuri\\
University of Exeter, UK\\
{\tt\small ac1151@exeter.ac.uk}
\and
Ayan Kumar Bhunia, Yi-Zhe Song, Anjan Dutta\\
Institute for People-Centred AI, University of Surrey, UK\\
{\tt\small \{a.bhunia, y.song, anjan.dutta\}@surrey.ac.uk}
}

\maketitle

\graphicspath{{figures/}}

\input{tex/00_abs.tex}

\input{tex/01_intro.tex}

\input{tex/02_related.tex}
\input{tex/03_method.tex}

\input{tex/04_expt.tex}

\input{tex/05_limitations.tex}

\input{tex/06_concl.tex}

\clearpage

{\small
\bibliographystyle{ieee_fullname}
\bibliography{egbib}
}

\end{document}


\title{Supplementary: Data-Free Sketch-Based Image Retrieval}



\author{Abhra Chaudhuri\\
University of Exeter, UK\\
{\tt\small ac1151@exeter.ac.uk}
\and
Ayan Kumar Bhunia, Yi-Zhe Song, Anjan Dutta\\
Institute for People-Centred AI, University of Surrey, UK\\
{\tt\small \{a.bhunia, y.song, anjan.dutta\}@surrey.ac.uk}
}

\maketitle

\graphicspath{{figures/}}

\section{Experimental Details}
\myparagraph{Datasets} The Sketchy dataset contains 75,471 sketches and 12,500 photos, with 60,502 additional photos from ImageNet \cite{Deng2009ImageNetAL} in its extension \cite{Liu2017DSH},
evenly distributed over 125 classes, with both class and instance level correspondences. The TU-Berlin dataset contains 20,000 sketches evenly distributed over 250 categories, with a total of 204,489 class-level natural image correspondences in its extension \cite{Liu2017DSH}.
The QuickDraw-Extended dataset contains 330,000 sketches and 204,000 photos from 110 categories.
Following existing literature \cite{Liu2017DSH, federici2020}, we use 10 and 50 randomly selected sketches per category from TU-Berlin and Sketchy respectively for testing the trained encoders in the category level SBIR setting, with the remaining sketches and photos used for training the teacher classifiers. We follow the same procedure as that of TU-Berlin for experimenting with the QuickDraw-Extended dataset.

\myparagraph{Pre-Trained Classifiers} We trained ResNet50 \cite{He2016DeepRL} models on the train splits of Sketchy and TU-Berlin to obtain the photo and sketch classifier networks that would act as teachers. We used Adam as the optimizer with a learning rate of 0.01 under an exponential decay rate of 0.98, and weight decay of $10^{-5}$. The photo and sketch classifiers were trained up to accuracies of 96.34\% and 93.81\% for Sketchy, and 92.70\% and 81.35\% for TU-Berlin respectively.

\myparagraph{Implementation Details}
%
Our estimator networks follow the architecture of the StyleGAN 2 \cite{Karras2020StyleGAN2} generator, trained using the Adam optimizer with a learning rate of 0.02. Our encoders have a ResNet50 \cite{He2016DeepRL} backbone, also trained using the Adam optimizer with a learning rate of $2 \times 10^{-3}$, decayed using a cosine annealing schedule. We initially train the estimators for 100 epochs in a warm-up phase for the estimated samples to stabilize and approach closer to the ones belonging to the true distribution. Thereafter, we train both the estimator and the generator pairs in an alternating manner (each frozen while the other is updated) for 500 epochs. In each epoch, we generate 10,000 positive pairs of photo-sketch reconstructions.

\myparagraph{Platform Details} We implement our data-free SBIR pipeline on an Ubuntu 20.04 workstation with a single NVIDIA RTX 3090 GPU, an 8-core Intel Xeon processor and 32 GBs of RAM, using the PyTorch \cite{Paszke2017PyTorch} deep learning framework. By the virtue of using a fixed-size, gradient-free queue for storing negative instances for contrastive learning of the encoders, our method bypasses dependencies on the batch-size, thus allowing us to perform the training end-to-end on a single GPU.



\subsection{Additional Details on Baselines}

\myparagraph{Sampling from a Gaussian Prior} The input photos and sketches constitute samples drawn from an $N \times N$ Gaussian distribution, where $N$ is the expected input spatial dimension for the downstream encoder. We assign labels to such samples in a manner that ensures equal number of samples across all classes. We then use these samples for training the encoders.

\myparagraph{Averaging Weights}
We speculate that averaging and using a single network for encoding photos and sketches helps bridge the modality gap.
As has been demonstrated time and again in relevant literature, modality-specific, semantically irrelevant features is the single biggest source of error in SBIR. With averaging weights, we are able to use a single network for encoding photos and sketches, while incorporating the knowledge about modality specific variances learned by the individual networks.

\myparagraph{Meta-Data Based Reconstruction} Following \cite{lopes2017DFKD}, we retain the means and the covariances of activations from all layers of the classifier. We then use them as metadata for input reconstruction, by generating samples that induce similar activation statistics across all layers of the classifier.


\section{Qualitative Ablations}

\subsection{Class Alignment}

\begin{figure*}[!ht]
    \centering
    \includegraphics[width=\textwidth]{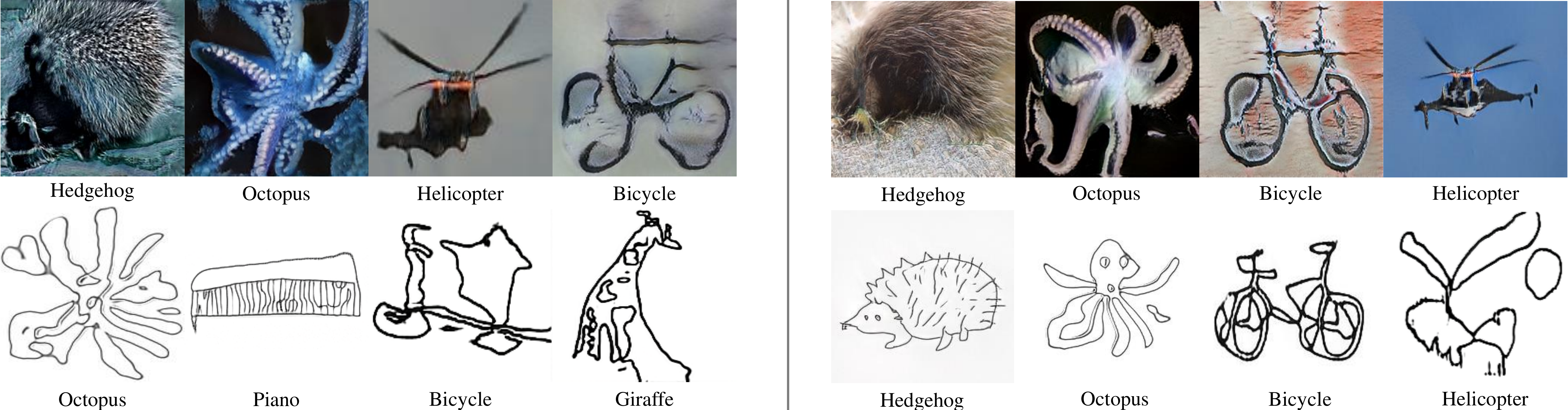}
    \caption{Photo and sketch reconstructions without (left) and with (right) the Class-Alignment loss ($\mathcal{L}_
    \text{align}$). Each column corresponds to a single reconstruction step using a common input noise vector $\xi$ fed in to the photo and sketch estimators respectively.}
    \label{fig:qualitative_ablation_class-alignment}
\end{figure*}

\begin{figure}
    \centering
    \includegraphics[width=\columnwidth]{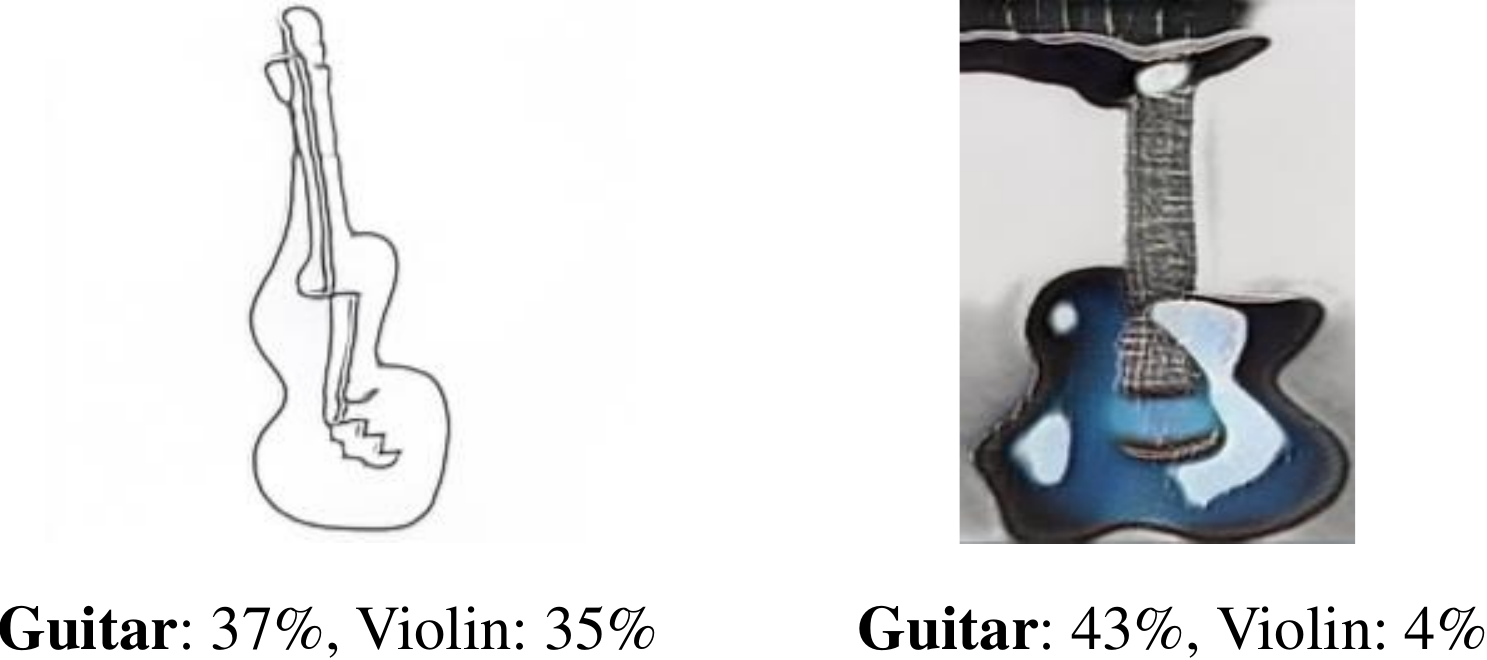}
    \caption{Examples of reconstructions for which the output distributions of the sketch and photo classifiers differ significantly.}
    \label{fig:differing_distributions}
\end{figure}

The contribution of the Class-Alignment loss ($\mathcal{L}_\text{align}$) in reconstructing semantically matching photo-sketch pairs given a common noise vector $\xi$ is visualized in \cref{fig:qualitative_ablation_class-alignment}. It can be seen that without $\mathcal{L}_\text{align}$, the estimators often produce samples that belong to largely different classes. This sends wrong signals to the downstream encoders in terms of learning a semantically meaningful metric-space. With the class-alignment loss, the estimators can be guaranteed to receive photo-sketch pairs that have the same class information, and hence qualify as correct positive pairs for optimizing the encoders.

Also, as discussed in Section 3.1 of the main text, pairing solely on the basis of discrete labels (obtained via an argmax on the teachers' output) may not faithfully represent the semantic content in a reconstruction. \cref{fig:differing_distributions} shows such an example. Even though their hard-labels are the same (Guitar), the distribution of class information is vastly different in the two images. Thus, for such pairs, it is important for their predicted distributions to be properly aligned (via an objective like $\mathcal{L}_\text{align}$) before they can actually be considered as positive pairs for downstream metric learning.

\subsection{Modality Guidance}

\begin{figure}
    \centering
    \includegraphics[width=\columnwidth]{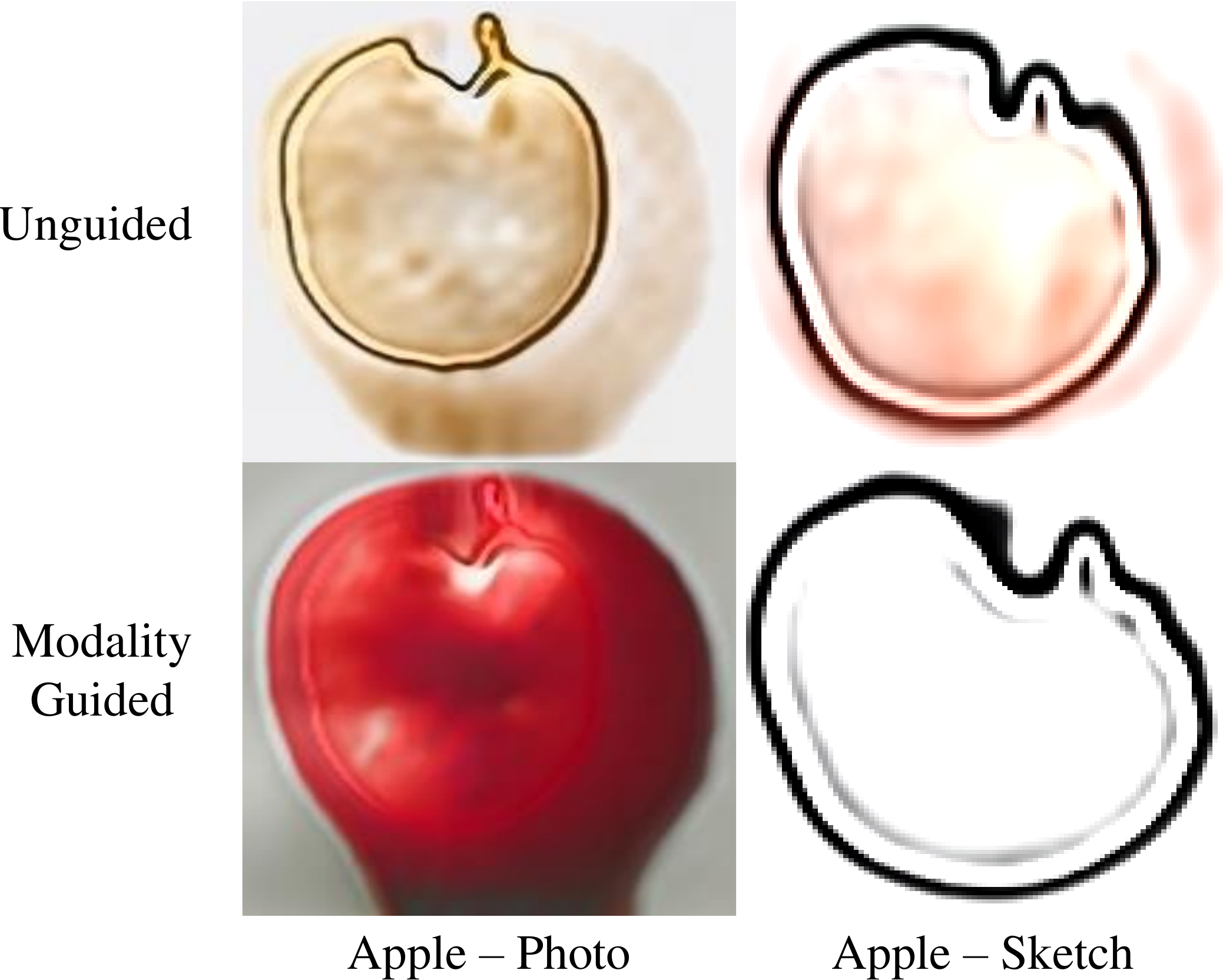}
    \caption{Reconstructed photos and sketches of an Apple in the presence and absence of the Modality Guidance loss ($\mathcal{L}_{\text{MG}}$).}
    \label{fig:qualitative_ablation_modality_guidance}
\end{figure}

\cref{fig:qualitative_ablation_modality_guidance} qualitatively demonstrates the contribution of our Modality Guidance Network ($d_\phi$), and its objective function $\mathcal{L}_\text{modal}$. It can be seen that Unguided estimators cannot make a clear distinction among the modalitites -- photo reconstructions contain object outlines like sketches, and sketch reconstructions contain colors. This happens because in presence of the class-alignment loss ($\mathcal{L}_\text{align}$), the estimators exchange information across modalities. Under such a circumstance, the estimators can minimize both semantic distance, as well as modality distance in order to minimize $\mathcal{L}_\text{align}$. The task of our Modality Guidance Network is to ensure that the estimators only minimize $\mathcal{L}_\text{align}$ by minimizing semantic distance, and not through the exchange of modality-specific information. With this, the Modality Guided estimators are bounded in their output space, producing clean and realistic reconstructions.

\begin{figure*}[!ht]
    \centering
    \includegraphics[width=\textwidth]{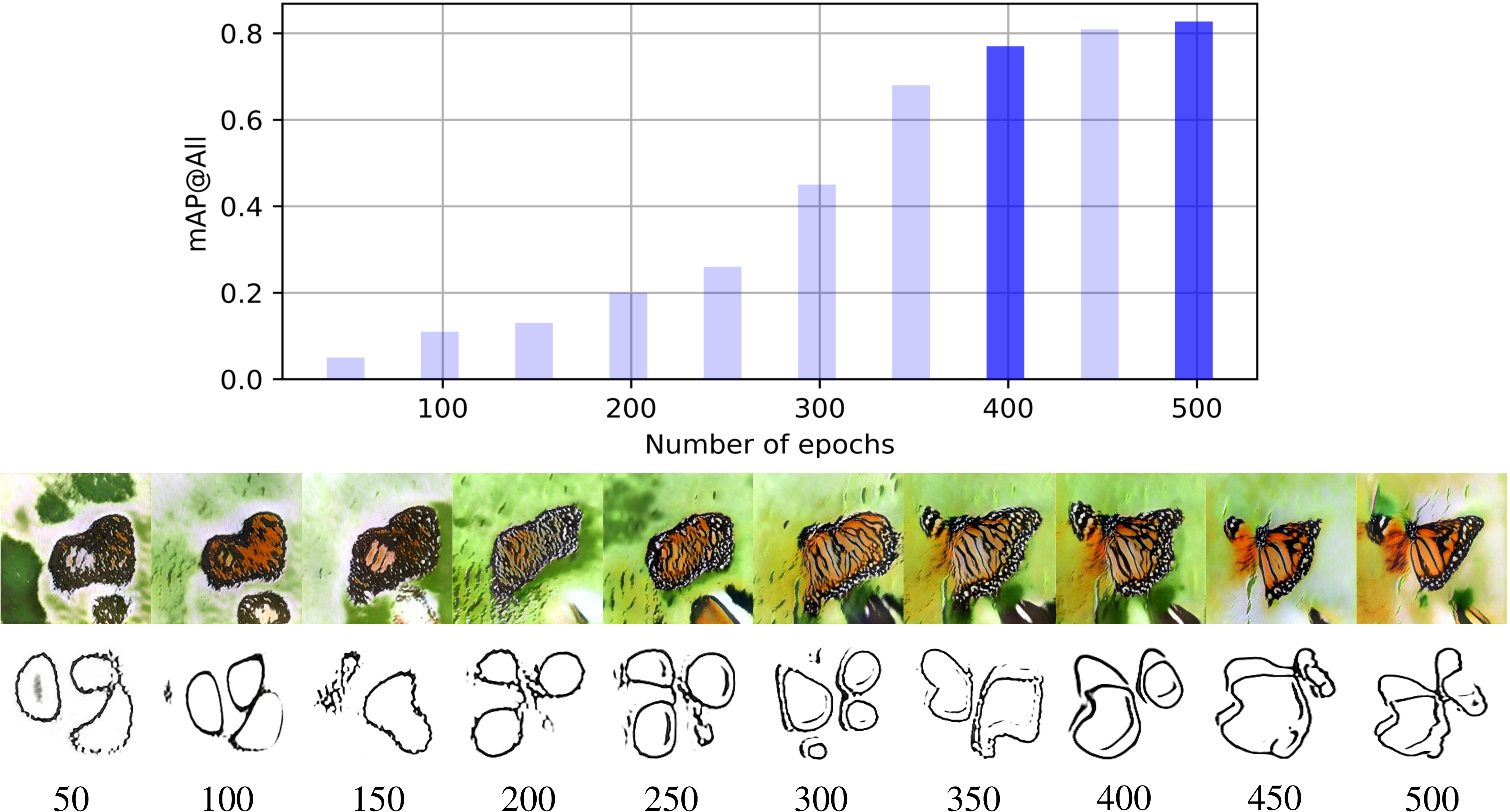}
    \caption{Variation in mAP@all, as well as reconstruction quality across epochs.}
    \label{fig:acc_vs_quality}
\end{figure*}

\subsection{Metric-Agnostic Adversarial Estimation}

\begin{figure}
    \centering
    \includegraphics[width=\columnwidth]{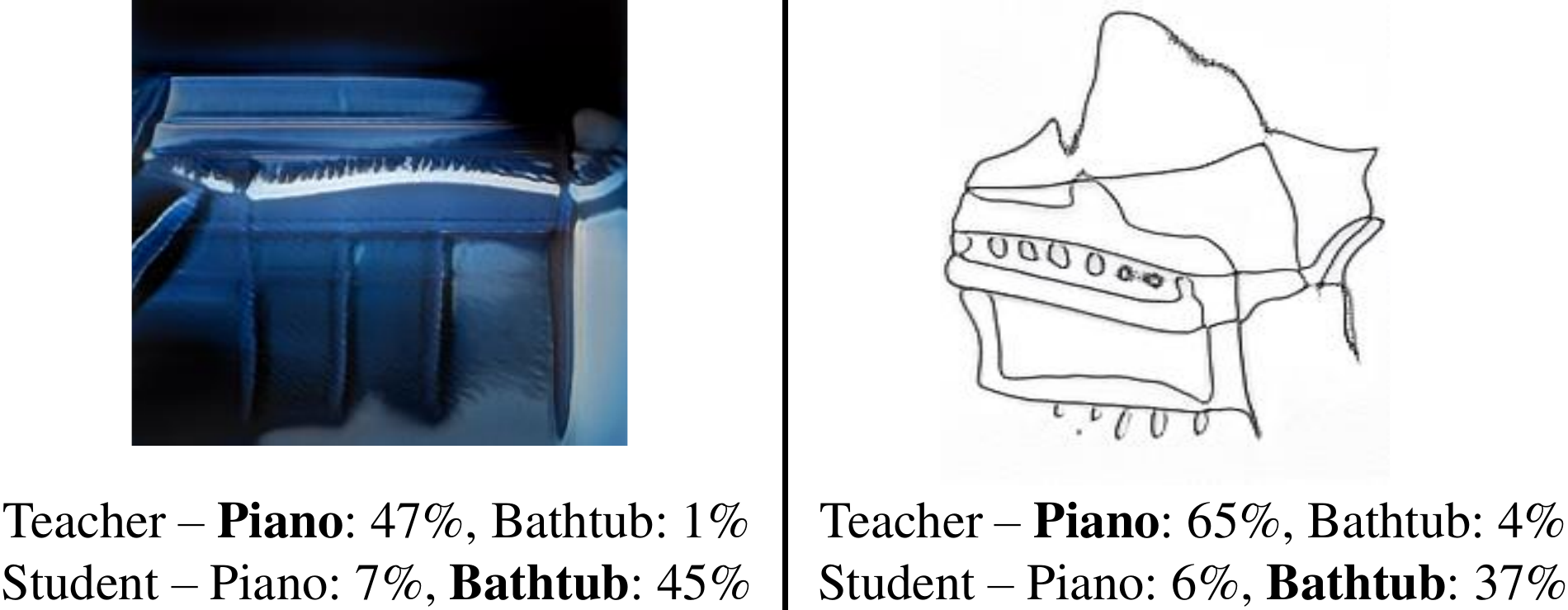}
    \caption{Sample reconstructions obtained by using our Metric-Agnostic Adversarial Estimation ($\mathcal{L}_{\text{adv}}$) criterion, with respective class-scores assigned by the teacher and the student.}
    \label{fig:qualitative_ablation_adversarial_estimation}
\end{figure}

\cref{fig:qualitative_ablation_adversarial_estimation} shows sample reconstructions obtained by optimizing our Metric-Agnostic Adversarial Estimation $\mathcal{L}_\text{adv}$ loss. While the teachers predict them to be instances of `Piano' with very high confidence while assigning the class of `Bathtub' a low probability, the predictions from the students is just of the opposite nature. This makes such samples the hardest for the student to encode, as their predictions are highly divergent from those of the teachers. Optimizing on such samples thus makes the student more robust to challenging real-world test cases.

\section{Reconstruction Quality and Performance}

\cref{fig:acc_vs_quality} shows the relationship between the quality of the reconstructed samples and the mean average precision (mAP) of the encoders on the Sketchy datasets. Training the estimators for longer helps in the reconstruction of more realistic samples. However, beyond a certain point, realism does not seem to significantly affect the retrieval performance. As the estimators start capturing the fundamental shape and texture of the object, there is a significant improvement in mAP from 0.45 (epoch 300) to 0.68 (epoch 350). With a few more epochs of training, the accuracy steadily increases to 0.77 (epoch 400). Beyond this point, the increase is much slower, although the quality of the reconstructions keep getting better.

\section{Training with Partial Class Overlap}

For the classes that are unknown to the photo classifier, we randomly initialize \emph{trainable} proxy-vectors to act as representatives for those classes, and concatenate them to the final layer neurons of the classifier. We do the same for the sketch classifier, and reorder the proxy vectors in both modalities to have consistency in class indices across modalities. For each reconstruction, if the sketch belongs to a class that is unknown to the photo classifier, we consider the prediction from the sketch classifier to be the ground-truth of the corresponding photo. We update the trainable proxies in the final layer of the photo classifier based on this information. We do a symmetric operation for the sketch classifier as well. Note that such alterations to the teachers are only possible because the reconstructed photos and sketches have been semantically aligned by our $\mathcal{L}_\text{align}$ objective. The rest of the process is performed as usual.

\clearpage

{\small
\bibliographystyle{ieee_fullname}
\bibliography{supp}
}

%% file: tex/00_abs.tex
\begin{abstract}
Rising concerns about privacy and anonymity preservation of deep learning models have facilitated research in data-free learning (DFL).
For the first time, we identify that for data-scarce tasks like Sketch-Based Image Retrieval (SBIR), where the difficulty in acquiring paired photos and hand-drawn sketches limits data-dependent cross-modal learning algorithms, DFL can prove to be a much more practical paradigm. We thus propose Data-Free (DF)-SBIR, where, unlike existing DFL problems, pre-trained, single-modality classification models have to be leveraged to learn a cross-modal metric-space for retrieval without access to any training data. The widespread availability of pre-trained classification models, along with the difficulty in acquiring paired photo-sketch datasets for SBIR justify the practicality of this setting. We present a methodology for DF-SBIR, which can leverage knowledge from models independently trained to perform classification on photos and sketches.
We 
evaluate our model on the Sketchy, TU-Berlin, and QuickDraw benchmarks, designing a variety of baselines based on state-of-the-art DFL literature, and observe that our method surpasses all of them by significant margins. Our method also achieves mAPs competitive with data-dependent approaches, all the while requiring no training data. Implementation is available at \url{https://github.com/abhrac/data-free-sbir}.
\end{abstract}

%% file: tex/01_intro.tex
\section{Introduction}

\begin{figure}
    \centering
    \includegraphics[width=\columnwidth]{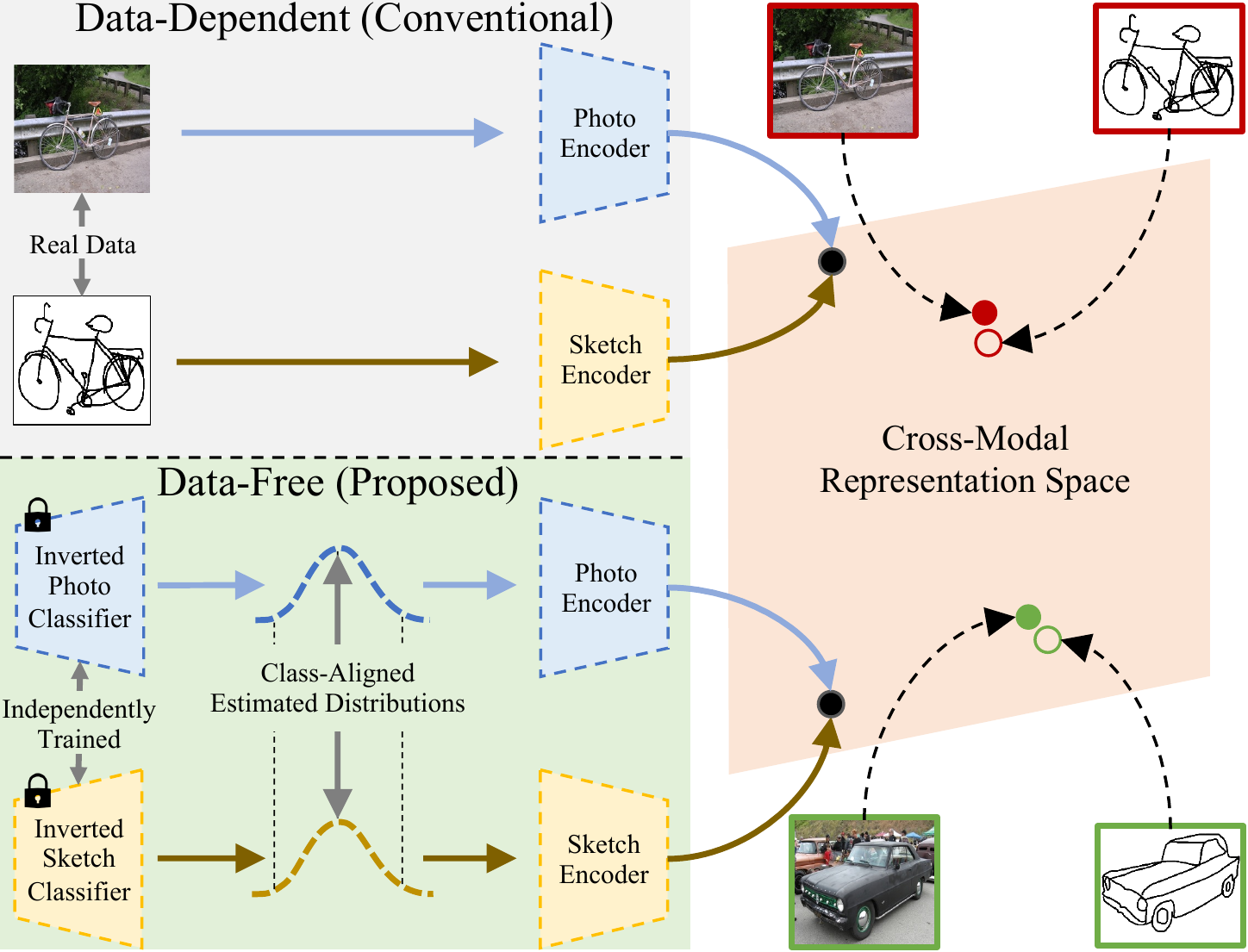}
    \caption{Our proposed Data-Free setting for SBIR does not need a real-world dataset of paired sketches and photos. Using only independently trained, modality-specific classifiers, it can estimate their train set distributions, as well as pair them at class-level for training the sketch and photo encoders.}
    \label{fig:teaser}
\end{figure}

%
Motivated by the high degree of expressiveness and flexibility provided by sketches, sketch-based image retrieval (SBIR) has emerged as a popular area of computer vision research \cite{Dutta2020SEMPCYCAny,Dey2019CVPR,Bhunia2022NoiseTolerantSBIR,Bhunia_2021_CVPR_MPaA,Chaudhuri2022XModalViT}.
SBIR is generally achieved by training photo and sketch encoders to respectively map photo and sketch inputs to a class or instance aligned common space. Training deep neural photo-sketch encoders for this task, however, requires datasets with matching photo-sketch pairs \cite{Sangkloy2016Sketchy, Yu2016SketchMe}. Unlike photos, sketches are fundamentally difficult to acquire as drawing them involve long time periods of laborious human participation. Driven by this practical constraint, the problem has been studied under a variety of data-scarce settings like semi-supervised \cite{Bhunia_2021_CVPR_MPaA}, few-shot class incremental \cite{Bhunia2022FewShotInc}, zero-shot \cite{Dutta2019CVPR}, and any-shot \cite{Dutta2020SEMPCYCAny}.
%
%
However, all such settings assume the availability of \emph{some} amount of instance/class-aligned data for training the encoders. With the tremendous effort involved in acquiring such labelled photo-sketch pairs \cite{Dey2019CVPR, Dutta2019CVPR, Bhunia_2021_CVPR_MPaA}, as well as the rising concerns about privacy, security and anonymity preservation abilities of deep learning models \cite{Chen2019DAFL, Lu2022April, Wang2022DST, Sanyal2022dfModelStealing}, such assumptions may no longer be practical.

With this view, we propose Data-Free Sketch-Based Image Retrieval (DF-SBIR), a novel setting that requires training photo and sketch encoders for retrieval, but with \emph{no} 
training data. Specifically, we only assume access to pre-trained photo and sketch classification models. In contrast to unsupervised cross-domain image retrieval \cite{hu2022feature} which only requires access to training data, but with no in-domain or cross-domain labels, our setting goes a step further and assumes access to no training data at all.
%
Since classification does not require cross-modal pairings as in SBIR, and owing to the recent advances in domain generalization \cite{Liu2022CNNFor2020s, Wortsman2022ModelSoups}, such pre-trained classifiers are widely available \cite{PretrainedModelsPyTorch, Wortsman2022ModelSoups}, making our setting quite practical.
%
Under this scenario, the problem of training the encoders can be posed in the light of data-free knowledge distillation (DFKD) \cite{lopes2017DFKD, Chen2019DAFL}. Classical knowledge distillation \cite{hinton2015distilling} aims to transfer knowledge from a pre-trained model (teacher) to a different model (student), by aligning the predictions of the latter with the former on train set inputs. Differently, DFKD aims to achieve this knowledge transfer without access to any form of training data. The conventional approach to DFKD involves the following two steps -- (1) Reconstructing the train set distribution of the teacher; (2) Training the student network to match its predictions with that of the teacher, on samples from the reconstructed distribution.
However, existing DFKD approaches so far have only been able to operate in a single modality, performing the same kind of task as that of the teacher. SBIR, being a cross-modal retrieval problem, cannot be tackled in the data-free setting by directly adapting the machineries developed for DFKD, the reasons for which we detail below.

\emph{First}, the teachers (being classifiers) and the students (being encoders) operate in metric spaces of different nature,
\ie, probabilistic and Euclidean respectively.
This restrains us from measuring the agreement between teachers and students in a straightforward way, thus preventing the direct application of state-of-the-art approaches from the DFKD literature like data-free adversarial distillation \cite{Micaelli2019ZeroShotKT, Choi2020dfAdv}. 
We address this by designing a unified, class-proxy based interface via which the teachers and students can interact. \emph{Second}, the sketch and the image classifiers that act as teachers are independently trained on modality-specific data. Their intermediate representations are thus modality sensitive. However, the representations learned by the encoders to be used for DF-SBIR need to be modality invariant. To this end, we introduce the concept of a modality guidance network, which constrains the reconstructions to belong to specific (photo/sketch) modalities. Training the encoders with such samples will ensure that they learn to eliminate unnecessary, modality-specific information. \emph{Third}, the independent training of the classifiers also mean that their train set distributions may not have direct class-level correspondence. To address this, we design our distribution estimation process to reconstruct class-aligned samples, \ie, ones that have class-level correspondence across the two modalities. This will guarantee the availability of matching photo-sketch pairs for the metric learning of the encoders. Our approach is hence able to perform Data-Free Learning Across Modalities and Metric-Spaces, which motivates us to abbreviate it as CrossX-DFL.

We make the following contributions -- (1) Propose Data-Free SBIR, a novel setting keeping in view the data-scarcity constraints arising from the collection of paired photos and sketches for SBIR, as well as concerns around privacy preservation; (2) A class-proxy based approach to perform data-free adversarial distillation from teachers with probabilistic outputs to students with outputs in the Euclidean space; (3) A novel technique to reconstruct class-aligned samples across independent modalities for cross-modal data-free knowledge distillation; (4) Introduce the concept of a modality guidance network to constrain the reconstructed sample distributions to specific modalities; (5) Extensive experiments on benchmark datasets and ablation studies that demonstrate the usefulness of our novel components in providing competitive performance relative to the data-dependent setting.

%% file: tex/02_related.tex
\section{Related Work}
\vspace{-0.2cm}
\keypoint{Data-Free Learning}
Data-free learning (DFL) \cite{Chen2019DAFL} aims to solve a learning problem without accessing any form of training data, possibly by leveraging the knowledge already learned by a different model, but for a related task. Such a setting provides practical benefits like data security \cite{Chen2019DAFL}, anonymity \cite{Lu_2022dfGeomReID}, privacy preservation \cite{lopes2017DFKD}, and model interpretability \cite{Mahendran2015DFInterpret} to name a few. The standard approach for performing DFL is based on the principle of model inversion \cite{Mahendran2015DFInterpret} -- given a pre-trained model (\emph{teacher}), the aim is to reconstruct its train set distribution by analyzing its activation patterns \cite{Chen2019DAFL, Yoo2019KnowledgeEW, Micaelli2019ZeroShotKT, lopes2017DFKD}. Instead of real data, this reconstructed distribution is then used to train a different downstream model (\emph{student}), thus making its learning process \emph{data-free}.
Such reconstructions can be ensured to have high fidelity and diversity by leveraging the teacher's batch normalization statistics \cite{Yin2020DreamingToDistill} and inducing contrastive teacher representations \cite{Fang2021dfContr} respectively. Along these lines, Micaelli \etal \cite{Micaelli2019ZeroShotKT} and Choi \etal \cite{Choi2020dfAdv} got a step further and introduced the idea of adversarial data-free distillation, which showed that training the downstream task networks (\emph{students}) specifically on the worst-case inputs (for which its predictions diverged maximally from the teacher) helped ensure robustness to factors of variation \cite{ben_tal2009RobustOptimization}. However, by its very design, the teacher and the student have to operate in the exact same representation space (either probabilistic \cite{Chen2019DAFL} or Euclidean \cite{Lu_2022dfGeomReID}), and its extension to cross-space distillation has remained yet unexplored. In this work, for the first time, we design a methodology that allows students to adversarially learn from a teacher, even when they operate in representation spaces of different nature.

\myparagraph{Applications of Data Free Learning}
Originally, DFL was proposed with the view of performing knowledge distillation (KD) \cite{hinton2015distilling} in a data-free manner \cite{lopes2017DFKD, Chen2019DAFL}. This was motivated by the fact that traditional KD required access to the teacher's train set \cite{Xie_2020_CVPR, Tian2020Contrastive, Park2019RelationalKD, Roth2021S2SD}, while rising concerns about privacy and anonymity preservation in deep learning models brought the practicality of such a requirement under question \cite{Chen2019DAFL, Lu_2022dfGeomReID}.
However, the potential of the underlying idea has helped DFL expand much beyond the horizons of KD, with its current applications ranging all the way from incremental person re-ID \cite{Lu_2022dfGeomReID}, image super-resolution \cite{Zhang2021DFSuperRes} and federated learning \cite{Zhang2022dfFedLearn}, to studying the security robustness of a model via black-box adversarial attacks \cite{Zhang_2022dfAttack, Wang2022DST} or model stealing \cite{Sanyal2022dfModelStealing}. To address the non-IID nature of local data distributions at the individual clients in the federated learning setting \cite{McMahan2017CommunicationEfficientLO}, principles of DFL \cite{Chen2019DAFL} allow local dataset reconstructions at the global level \cite{Zhang2022dfFedLearn}. Going beyond such applications, Lu \etal \cite{Lu2022April} has showed that using online gradient update based reconstructions can be used for exposing privacy preservation gaps in federated learning systems. The latter idea of uncovering a model's security deficits via DFL has further been studied in the context of black-box adversarial attacks \cite{Cheng2019ImprovingBA} for designing a data-free generator that maximizes information leakage from the target model to the substitute model \cite{Zhang_2022dfAttack}. The same approach has also been applied in model stealing \cite{Sanyal2022dfModelStealing}, or for learning a single dynamic substitute network for multiple target models with different downstream tasks \cite{Wang2022DST}. However, none of the above works deal with cross-modal retrieval tasks (like SBIR), where generating paired instances across modalities is a challenging and time-intensive process. We address this issue through our novel construction of semantically-aligned, modality-specific estimators.

\myparagraph{Sketch-Based Image Retrieval}
SBIR methods relying on hand-crafted features focused on matching geometrically similar substructures via gradient field HOG \cite{Hu2013GFHOG}, deformable parts model \cite{DeformableParts}, histogram of edge local orientations \cite{Saavedra2014SHELO}, learned key shapes \cite{Saavedra2015LKS}. To address the sensitivity of such methods to the large sketch-photo modality gap, deep SBIR models employing a deep triplet network to learn a common embedding space were proposed in \cite{Yu2016SketchMe,Sangkloy2016Sketchy}, which was extended in \cite{FgsbirSpatialAttention} by the introduction of spatial attention, and in \cite{PangKaiyue2020SMJP, bhunia2021sketch2vec} via self-supervised pretext tasks. Generative models have also shown promising results in SBIR, employing ideas like disentangling style and semantic contents \cite{Sain_2021_CVPR_StyleMeUp,BMVC2017_46}, cross-domain image synthesis \cite{BMVC2017_46}, and reinforcement learning based sketch generation \cite{Bhunia_2021_CVPR_MPaA}. All the above methods rely on expensive continuous valued distance computation, hampering scalability, which \cite{Liu2017DSH} addressed by learning binary hash codes for sketches. Explainability \cite{Alaniz22SketchPrimitives} and robustness \cite{federici2020, Bhunia2022NoiseTolerantSBIR} have also become recent areas of focus in SBIR research.
The data scarcity issue in SBIR stemming from the difficulty in acquiring hand-drawn sketches has been addressed by studying the problem in semi-supervised \cite{Bhunia_2021_CVPR_MPaA} and zero-shot settings \cite{Yelamarthi2018ECCV, Dutta2019CVPR, Dey2019CVPR}. However, performing SBIR in a scenario where one has access to absolutely \emph{no} training data, has never been explored before. 
With the widespread availability of pre-trained classification models on benchmark datasets \cite{PretrainedModelsPyTorch, Wortsman2022ModelSoups}, and the costs and constraints involved in acquiring datasets with paired photo-sketch instances, data-free SBIR emerges as a promising research direction.

%% file: tex/03_method.tex
\begin{figure*}
    \centering
    \includegraphics[height=6.5cm,width=\textwidth]{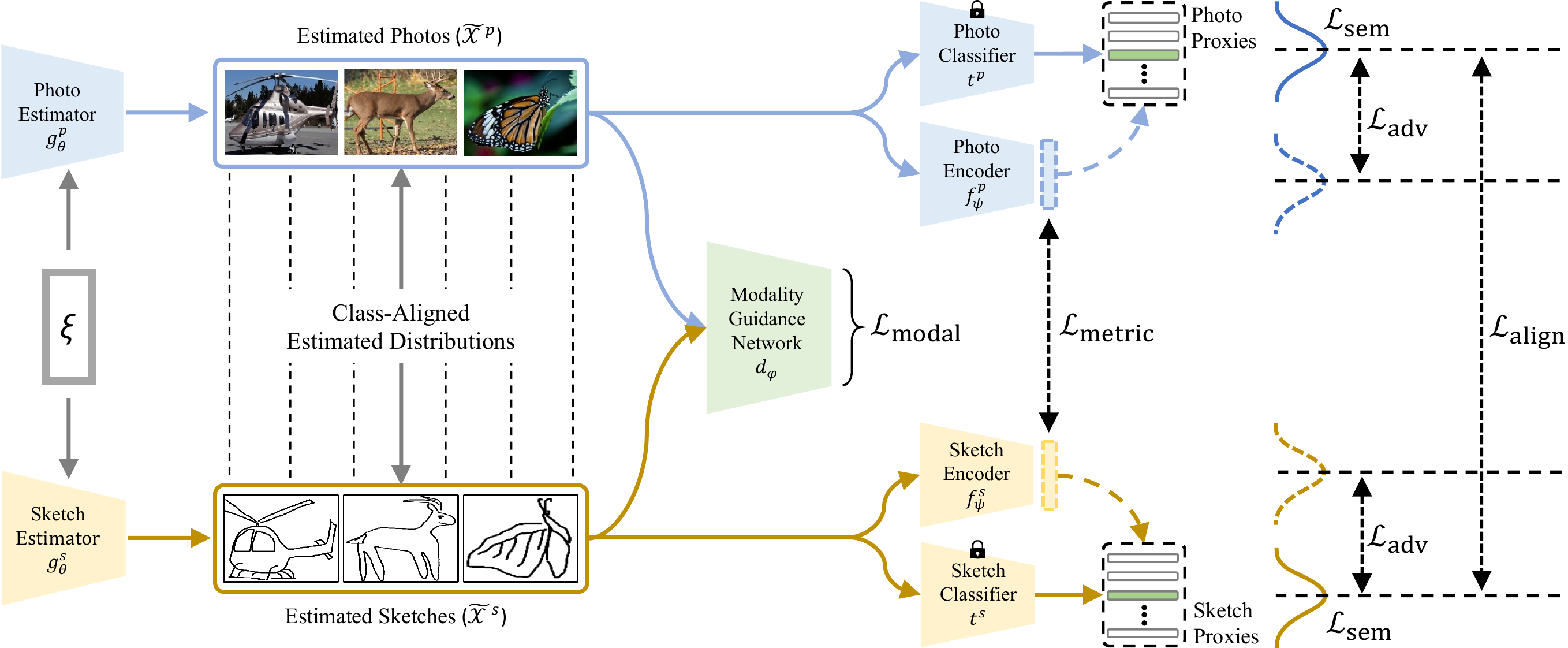}
    \caption{Our CrossX-DFL training pipeline for Data-Free SBIR -- The photo and sketch estimators reconstruct the train set distributions of their corresponding classifiers. The estimation process is ensured to have semantic consistency and class-level, instance-wise correspondence via $\mathcal{L}_\text{sem}$ and $\mathcal{L}_\text{align}$, while $\mathcal{L}_\text{modal}$ constrains the estimated distributions to belong to specific modalities. $\mathcal{L}_\text{adv}$ incentivises the estimators to produce boundary samples that are the hardest for the encoders to correctly encode. Optimizing the performance of the encoders on such samples ensures its robustness to semantic variations. The reconstructed samples from the estimated photo and sketch distributions, \ie, $\Tilde{\mathcal{X}}^p$ and $\Tilde{\mathcal{X}}^s$ respectively, are used for training the encoders in a data-free manner by minimizing $\mathcal{L}_\text{metric}$.}
    \label{fig:model_diagram}
\end{figure*}

\section{CrossX-DFL}
\myparagraph{Notations}
Let $t^p$ and $t^s$ be photo and sketch classifiers (teachers) already trained over input distributions $\mathcal{X}^p$ and $\mathcal{X}^s$ respectively. Let $g^p_\theta$ and $g^s_\theta$ respectively be photo and sketch estimator networks that learn to map random Gaussian noise vectors $\xi \in [0, 1]^n$ to elements in the training domains of the photo and sketch classifiers.
We denote by $f^p_\psi$ and $f^s_\psi$, the photo and sketch encoders (students) that are trained to compute metric space embeddings for photos and sketches respectively, and can thus be used for the task of SBIR.
For brevity, we will omit the parameters and the modality superscripts while referring to the networks in generic scenarios (\eg, $f^p_\psi$ will be denoted as $f$).


\myparagraph{Overview}
We approach the task of data-free SBIR by designing a bespoke process for performing data-free knowledge distillation from the classifiers to the encoders. The process consists of estimating the train set distribution of the classifiers followed by leveraging that estimation to train downstream photo and sketch encoders for retrieval. The steps are summarized below:\\
\noindent \textbf{Step 1:} Estimating the Input Distribution -- We train the estimator networks $g$ to produce approximations of the train set distributions of the classifiers $t$. Based on feedbacks from the classifiers $t$ and the encoders $f$, the estimators generate distributions $\Tilde{\mathcal{X}}$ such that they are as close to the true sample distributions $\mathcal{X}$ as possible.
Since the classifiers act as teacher networks, their weights are kept frozen throughout the process.
\\
\noindent \textbf{Step 2:} Training the Encoders -- The encoders $f$ are then trained to produce metric space embeddings for the approximated training distributions $\Tilde{\mathcal{X}}$ such that matching photo-sketch pairs $(\Tilde{\vectorname{x}}^p, \Tilde{\vectorname{x}}^s) \in (\Tilde{\mathcal{X}}^p, \Tilde{\mathcal{X}}^s)$ are mapped close to each other in the representation space, and non-matching pairs are mapped farther apart.\\
\noindent \textbf{Step 3:} Retrieval -- Given the trained sketch encoder $f^s_\psi$ and photo encoder $f^p_\psi$, a gallery of photos $\{\vectorname{x}^g_1, \vectorname{x}^g_2, ..., \vectorname{x}^g_k\}$ and a query sketch $\vectorname{x}^s$, we return a ranking over the gallery such that $\vectorname{x}^g_i$ is ranked above $\vectorname{x}^g_j$ \emph{iff}:\\
\begin{equation*}
    f^s(\vectorname{x}^s) \cdot f^p(\vectorname{x}^g_i) \geq f^s(\vectorname{x}^s) \cdot f^p(\vectorname{x}^g_j)
\end{equation*}
%
Step 3 is generic across all SBIR methods and is common to both data-free and data-driven settings. We further elaborate on Steps-1 and 2, which are specific to the data-free setting, below in \cref{sec:estimating_distribution,sec:training_encoders} respectively.

\subsection{Estimating the Input Distribution}
\label{sec:estimating_distribution}
The primary constraint that the reconstructed samples of the estimated distributions need to satisfy is that their class and modality memberships must be unambiguous.
Since the downstream encoders need positively paired photos and sketches for contrastive learning, there needs to be a constraint enforcing the semantic alignment of the two distributions.
One could also ensure the robustness of the encoders by presenting it only with the samples that are the hardest for it to correctly represent. Based on these requirements, we formulate the input distribution estimation of the two classifiers as follows.

\myparagraph{Semantic Consistency}
To keep the reconstructed elements of the estimated distribution semantically consistent, we constrain the estimation process to minimize the entropy of the classifier's output distribution, while maximizing the activation values of the representation layer. For a reconstructed input $\Tilde{x} = g(\xi)$
and its corresponding predicted probability distribution $\hat{Y} = t(\Tilde{x}) = [\hat{y}_1, \hat{y}_2, ..., \hat{y}_C]$ obtained from the classifier $t$ over a set of $C$ classes, we minimize the following:
\begin{align*}
    \mathcal{L}_\text{sem} = \sum_{i = 1}^C y_i \log \hat{y}_i - |t_{n-1}(\Tilde{x})|,
\end{align*}
where $y_i = 1$ if $\max \hat{Y} = \hat{y}_i$, or 0 otherwise, and $t_{n-1}(\Tilde{x})$ is the output of the teacher's representation layer. Minimizing the classification entropy helps to ensure that the reconstructions $\Tilde{x}$ come from distinct classes\footnote{We ensure that set of estimated samples are class-balanced via the information entropy criterion as in \cite{Chen2019DAFL}.}
that the classifier can recognize, while maximizing the magnitude of the classifiers' representation layer activations incentivizes the estimators to introduce meaningful contents in the reconstructions, as observed in \cite{Chen2019DAFL}.

\myparagraph{Class Alignment}
Even though $\mathcal{L}_\text{sem}$ incentivizes meaningful class memberships, it does not guarantee that the corresponding samples between the two distributions would belong to the same class. Also, pairing samples solely on the basis of discrete class labels does not guarantee exact correspondence. This is because the degree/distribution of class information between the two samples may significantly differ, even if the class label happens to be the same.
Hence, we direct the estimation process in a manner such that the corresponding samples across the reconstructed distributions are semantically-aligned. This ensures more fine-grained correspondence between the reconstructed sketch-photo pairs.
For a given noise vector $\xi$, we enforce the photo and the sketch estimators to map it to their corresponding modality specific reconstructions, but in a way such that they represent elements of the same class. We achieve this by minimizing the symmetrized Kullback-Leibler (KL)-divergence between the output distributions of the photo and the sketch classifiers as follows:
\begin{align*}
    \mathcal{L}_\text{align} = \frac{1}{2}\sum_{i=1}^C y^p_i \log \frac{y^p_i}{y^s_i} + y^s_i \log \frac{y^s_i}{y^p_i},
\end{align*}
where $y^p_i$ and $y^s_i$ respectively stand for the probabilistic scores predicted by the photo and sketch classifiers for the $i$-th class.

\myparagraph{Modality Guidance} We ensure that the estimators produce samples that are specific to their corresponding modalities by incorporating feedback from a modality guidance network. This guarantees that the reconstructions have sufficient modality-specific information, which the encoder needs to be able to eliminate in order to learn robust representations for retrieval. The modality guidance network $d$ is a classifier parameterized by $\phi$ that discriminates between photos and sketches. The modality guidance loss is formulated as follows:
\begin{align*}
    \mathcal{L}_\text{modal} = - m \log d(\Tilde{x}) - (1 - m) \log (1 - d(\Tilde{x})),
\end{align*}
where $m = 1$ if $\Tilde{x}$ comes from the photo estimator, and 0 if it comes from the sketch estimator.
We independently optimize the weights of the estimators and the modality guidance network to minimize $\mathcal{L}_\text{modal}$. The convergence of this criterion
will imply that the estimators are able to incorporate modality-specific information into the reconstructions $\Tilde{x}$, since it is the only premise on which $d_\phi$ can distinguish the two modalities (as they are already semantically-aligned as ensured by minimizing $\mathcal{L}_\text{sem}$ and $\mathcal{L}_\text{align}$).

\myparagraph{Metric-Agnostic Adversarial Estimation} Inspired by robust optimization \cite{ben_tal2009RobustOptimization}, existing literature on data-free knowledge distillation \cite{Micaelli2019ZeroShotKT, Choi2020dfAdv} has shown that the efficiency of the distillation process can be significantly improved by identifying the samples which are the hardest for the student to classify.
%
Concretely, this objective translates to tasking the estimators to produce boundary samples, \ie, the ones that maximize the \emph{disagreement} between the teachers (classifiers) and the students (encoders).
%
%
However, in data free SBIR, the teacher being a classifier outputs probability distributions in $[0, 1]^C$, while the student being an encoder produces representation vectors in $\mathbb{R}^n$, which restrains us from directly measuring the \emph{disagreement} between the teachers and the students in a straightforward way.
To address this issue, we treat the final layer neurons of the teacher (classifier) networks as class-proxies \cite{Movshovitz-Attias2017ProxyNCA} against which we perform metric learning with the students' representations. We thus adapt the proxies to provide an interface between the metric spaces of the teachers and the students, allowing outputs from both networks to be compared with each other.
Let $K^p = [\vectorname{k}^p_1, \vectorname{k}^p_2, ..., \vectorname{k}^p_C]$ and $K^s = [\vectorname{k}^s_1, \vectorname{k}^s_2, ..., \vectorname{k}^s_C]$ be the sets of class-proxies 
for the photo and the sketch modalities respectively (generically denoted as $K$). We align the metric spaces of the classifiers and the encoders by mapping representations obtained from the encoders $f(\Tilde{\vectorname{x}})$ to the corresponding class proxy vectors $\vectorname{k} \in K$, minimizing the following Proxy Anchor loss \cite{Kim2020CVPR}:
\begin{align*}
    \mathcal{L}_\text{metric} = \frac{1}{|K^+|}\sum_{\vectorname{k} \in K^+} \log (1 + \sum_{\Tilde{x} \in \Tilde{X}^+_\vectorname{k}} e^{-\alpha(s(f(\Tilde{x}), \vectorname{k}) - \delta)}) + \\
    \frac{1}{|K|}\sum_{\vectorname{k} \in K} \log (1 + \sum_{\Tilde{x} \in \Tilde{X}^-_\vectorname{k}} e^{\alpha(s(f(\Tilde{x}), \vectorname{k}) + \delta)})
\end{align*}
%
where $\alpha$ and $\delta$ respectively denote the margin and scaling factors, $K^+$ denote the set of positive proxies corresponding to the the samples in a minibatch, $\Tilde{X}^+_\vectorname{k}$ and $\Tilde{X}^-_\vectorname{k}$ respectively denote the matching and non-matching samples for the proxy $\vectorname{k}$, $f^p(\Tilde{x}) = \Vec{0}$ if $\Tilde{x}$ is obtained from the sketch estimator, and $f^s(\Tilde{x}) = \Vec{0}$ if $\Tilde{x}$ is obtained from the photo estimator.

The estimators $g$ and encoders $f$ then adversarially optimize the following objective:
\begin{align*}
    \mathcal{L}_\text{adv} = \min_{f}\max_{g}\{\operatorname{D_{KL}}(\operatorname{softmax}(f(g(\xi)) \cdot K), \hat{Y})\},
\end{align*}
where $D_\text{KL}$ stands for KL divergence. The maxima of the $D_\text{KL}$ between the output distributions of the classifier and encoder corresponds to the samples that are the hardest for the encoders to correctly encode. By aiming for this maximum, the estimators learn to generate those hard samples,
while the encoders aim to optimize their performance on such samples by getting closer to the classifiers' predictions.

\subsection{Training the Encoders}
\label{sec:training_encoders}
As ensured by minimizing $\mathcal{L}_\text{align}$, the reconstructed distributions $\Tilde{\mathcal{X}}^p$ and $\Tilde{\mathcal{X}}^s$ are class-aligned, \ie, for each index $i$, $\Tilde{\vectorname{x}}^p = \Tilde{\mathcal{X}}^p(i)$ and  $\Tilde{\vectorname{x}}^s = \Tilde{\mathcal{X}}^s(i)$ belong to the same class. With such positive pairs available out of the box, we train the photo and the sketch encoders to minimize the following queue-based version of the InfoNCE loss \cite{Oord2018RepresentationLW}:
\begin{align*}
    \mathcal{L}_\text{enc} = -\log \frac{\exp{(f^s(\Tilde{\vectorname{x}}^s) \cdot f^p(\Tilde{\vectorname{x}}^p)/ \tau) }}{\displaystyle\sum_{\vectorname{z^p} \in \mathcal{Q}}\exp{(f^s(\Tilde{\vectorname{x}}^s) \cdot \vectorname{z}^p  / \tau)}},
\end{align*}
where $\mathcal{Q}$ is a gradient-free queue of photo sample representations from previous mini-batches and $\tau$ is a hyperparameter controlling the spread of the embedding distribution \cite{Wu2018CVPR, He2020MomentumCF}.

\myparagraph{Learning objective} We optimize the estimators $g$, encoders $f$, and the modality discriminator $d$ parameterized by $\theta, \psi$, and $\phi$ respectively via the following gradient updates:
\begin{align*}
    \theta &\xleftarrow{} \theta - \eta \nabla_\theta(\mathcal{L}_\text{sem} + \mathcal{L}_\text{align} + \mathcal{L}_\text{modal} + \mathcal{L}_\text{adv}), \\
    \psi &\xleftarrow{} \psi - \eta \nabla_\psi(\mathcal{L}_\text{metric} + \mathcal{L}_\text{adv} + \mathcal{L}_\text{enc}), \\
    \phi &\xleftarrow{} \phi - \eta \operatorname{\nabla_\phi} \mathcal{L}_\text{modal},
\end{align*}
where $\eta$ is the learning rate. Upon convergence, we only retain the encoders $f_\psi$ for performing retrieval.

%% file: tex/04_expt.tex
\section{Experiments}

\myparagraph{Datasets} We evaluate our model on the three most common large scale SBIR datasets, namely Sketchy \cite{Sangkloy2016Sketchy, Liu2017DSH}, TU-Berlin \cite{eitz2012TUBerlin, Liu2017DSH} and QuickDraw-Extended \cite{Dey2019CVPR}.
We provide details on their statistics, as well as train-test splits in the supplementary. For Sketchy, the sketches and photos have instance-level correspondences. However, TU-Berlin \cite{eitz2012TUBerlin} and QuickDraw \cite{Jongejan2016QuickDraw} were originally sketch-only datasets, which were extended for SBIR in \cite{Liu2017DSH} and \cite{Dey2019CVPR} by adding photos from ImageNet \cite{Deng2009ImageNetAL} and \emph{Flickr} respectively, which only provide class-level, but \emph{no} instance-level correspondence. Therefore, experiments on these two classes of datasets pose very different challenges for our proposed model to handle.
We use mean average precision (mAP@all) and precision at 200 (Prec@200) as evaluation metrics in accordance with the convention in existing works \cite{Liu2017DSH, federici2020}. In the supplementary, we additionally provide details on the pre-trained classifiers (teachers), hyperparameter choices, and our experimentation platform.

\begin{figure*}
    \centering
    \includegraphics[width=\textwidth]{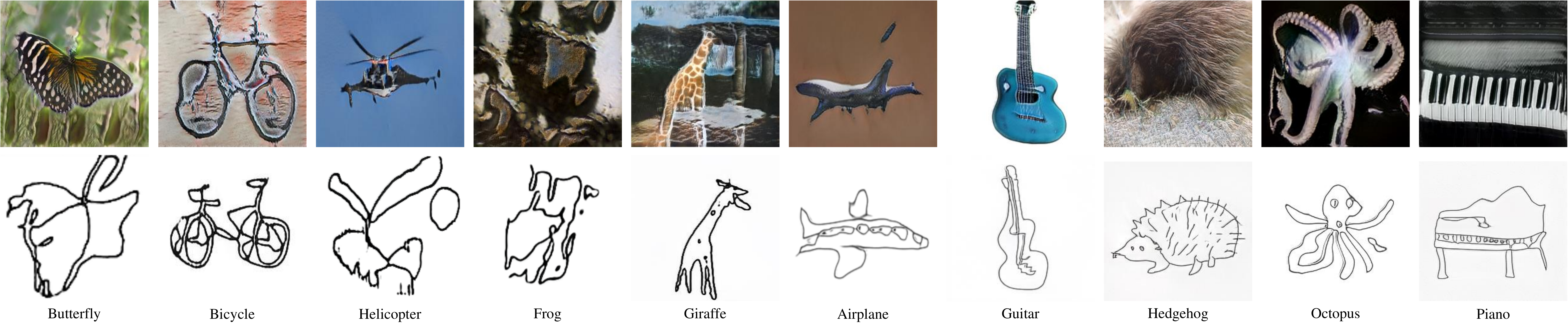}
    \caption{Data-free photo and sketch reconstructions of the Sketchy and TU-Berlin datasets produced by our estimator networks.}
    \label{fig:qualitative_results}
\end{figure*}

\subsection{Comparison with Baseline Approaches}
In \cref{tab:ComparisonBaselines}, we compare our method with various baseline approaches derived from existing state-of-the-art DFL literature. Below, we discuss our observations, with additional implementation details and analyses in the supplementary.

\begin{table}[!t]
    \centering
    \resizebox{\columnwidth}{!}{
    \begin{tabular}{l|c|c|c|c|c|c}
    \hline
    \multirow{2}{*}{\textbf{Method}} & \multicolumn{2}{c|}{\textbf{Sketchy}} & \multicolumn{2}{c|}{\textbf{TU-Berlin}} & \multicolumn{2}{c}{\textbf{QuickDraw}} \\
    & \textbf{mAP@all} & \textbf{Prec@200} & \textbf{mAP@all} & \textbf{Prec@200} & \textbf{mAP@all} & \textbf{Prec@200}\\
    \hline
    Classifier Only \cite{He2016DeepRL} & 0.530 & 0.542 & 0.330 & 0.338 & 0.160 & 0.180 \\
    Uni-Modal Distillation \cite{Chen2019DAFL} & 0.529 & 0.537 & 0.291 & 0.295 & 0.130 & 0.140 \\
    Gaussian Prior \cite{Chen2019DAFL} & 0.365 & 0.391 & 0.110 & 0.126 & 0.080 & 0.110 \\
    Averaging Weights \cite{Wortsman2022ModelSoups} & 0.625 & 0.630 & 0.450 & 0.473 & 0.300 & 0.320 \\
    Meta-Data \cite{lopes2017DFKD} & 0.573 & 0.576 & 0.380 & 0.395 & 0.200 & 0.221 \\
    Alternative Data \cite{Deng2009ImageNetAL, wang2019ImageNetSketch} & 0.656 & 0.680 & 0.510 & 0.530 & 0.290 & 0330 \\
    \hline
    \textbf{Ours (CrossX-DFL)} & \textbf{0.827} & \textbf{0.831} & \textbf{0.680} & \textbf{0.693} & \textbf{0.400} & \textbf{0.410} \\
    \hline
    \end{tabular}}
    \caption{Quantitative comparison of CrossX-DFL with baselines.}
    \label{tab:ComparisonBaselines}
\end{table}

\myparagraph{Classifier Representations} The pre-trained ResNet50 \cite{He2016DeepRL} classifiers are made to act as encoders. We remove the final classification layer and use the output from the representation layer of the classifiers to perform SBIR.
Since their representations spaces have not been aligned by a contrastive objective like $\mathcal{L}_\text{enc}$, directly using the classifiers as encoders result in sub-par retrieval performance.

\myparagraph{Uni-Modal Knowledge Distillation} Since all state-of-the-art data-free knowledge distillation approaches operate on uni-modal data \cite{lopes2017DFKD, Chen2019DAFL, Choi2020dfAdv, Fang2021dfContr}, we adapt them to train the photo and sketch students, but without any cross-modal interaction.
Thus, during the distillation phase, the photo and the sketch students learn to match their outputs with those of their modality-specific teachers for classification, but do not interact with each other via class-alignment or contrastive losses for optimizing retrieval performance. We then directly use the representations obtained from such uni-modal networks for SBIR.
Although promising, the performance of this approach has to be upper-bounded by what is achieved using the teachers directly. This phenomenon is also supported by our experimental results. Related literature suggests that it is possible to train students to surpass the performance of their teacher by making the former larger in size than the latter \cite{Xie_2020_CVPR}. However, we used same-sized backbones for both to ensure fair comparison.

\myparagraph{Sampling from a Gaussian Prior} Instead of estimating the train set distributions by inverting the classifiers, inspired by related approaches in data-free learning \cite{Chen2019DAFL}, we assume a Gaussian prior for their distributions.
Considering the pixels of the photos and sketches as independently and identically distributed Gaussians turns out to be too general an assumption. Identifying the semantic structures in such a manifold would require dense sampling of these distributions, which is computationally intractable. Hence, with a similar training time-scale as CrossX-DFL, this approach performs significantly worse.

\myparagraph{Averaging Weights}
Averaging the weights of all the networks in an ensemble has recently been shown to outperform even the best model in the ensemble, providing better robustness and domain generalization \cite{Wortsman2022ModelSoups}. In similar vein, we can pose SBIR as a domain generalization problem for the aggregated model.
We average the weights of the photo and sketch classifiers, remove the classification layer, and obtain a single network for encoding both photos and sketches.
This achieves around a 9\% improvement over the individual classifier. However, the performance gap with our CrossX-DFL still remains significant.

\myparagraph{Meta-Data Based Reconstruction}
We follow \cite{lopes2017DFKD} by using statistical metadata for obtaining photo/sketch reconstructions.
This provides better performance than the classifier-only setting. However, the reconstructions are not very efficient due to the lack of semantic consistency, modality specificity, sample hardness, etc.

\myparagraph{Training with Alternative Datasets}
Due to the limited availability of bespoke datasets for training SBIR models, we leverage more easily available datasets for related tasks like classification, domain generalization, robustness, etc. We use the photos from ImageNet \cite{Deng2009ImageNetAL} and their corresponding sketches from ImageNet-Sketch \cite{wang2019ImageNetSketch} to train the photo and sketch encoders respectively.
We use the class-level correspondence information from the two datasets to generate positive photo-sketch pairs for contrastive learning.
This setting gives the highest accuracy among all baselines. However, it still remains much lower than what is achieved by our CrossX-DFL. We speculate that although training on related datasets may help in learning task-specific (here retrieval) representations, the large error rate stems from the distribution shifts across different datasets. Our CrossX-DFL, by the virtue of the semantic consistency, class-alignment, modality guidance and metric-agnostic adversarial estimation criteria, is able to reconstruct a much more faithful approximation of the train set distribution, thereby providing significantly superior performance.

\begin{table}[!t]
    \centering
    \resizebox{\columnwidth}{!}{
    \begin{tabular}{l|c|c|c}
        \hline
        \textbf{Objective} & \textbf{Data-Dependent} & \textbf{Data-Free} & $\Delta$ \\
        \hline
        Siamese \cite{Chopra2005SiameseLoss} & 0.715 & 0.679 & 0.036 \\
        Triplet \cite{Sangkloy2016Sketchy} & 0.772 & 0.750 & 0.022 \\
        MIB \cite{federici2020} & 0.871 & 0.815 & 0.056 \\
        \textbf{Ours (CrossX-DFL)} \cite{Wu2018CVPR} & 0.862 & 0.827 & 0.035 \\
        \hline
    \end{tabular}}
    \caption{Data-Dependent \textit{vs.} Data-Free settings on Sketchy.}
    \label{tab:comparison_with_data_dependent}
\end{table}

\subsection{Comparison with the Data-Dependent Setting}
The upper-bound for the performance of our model is given by the mAP@all obtained using encoders directly trained on the original datasets. We empirically measure these upper-bounds for various commonly used, state-of-the-art contrastive learning objective functions, and analyze how close our algorithm gets to each of them. To this end, we train our encoders using both the original train set photos and sketches from Sketchy (data-dependent), as well the estimated distributions obtained by our method (data-free). We summarize our findings in \cref{tab:comparison_with_data_dependent}. For all objectives, we are able to get quite close to the data-dependent setting, while using \emph{no} training data at all. For instance, with the Triplet loss, our method achieves a difference ($\Delta$) of only 0.022 with the data-dependent setting. However, with Queued-InfoNCE, our method achieves the highest mAP@all of 0.827, with a $\Delta$ of 0.035 with the data-dependent setting, which is also quite competitive.

\subsection{Qualitative Results}

\cref{fig:qualitative_results} shows sample photo and sketch reconstructions produced by our estimator networks. We would like to emphasise that our end-goal is \emph{not} to generate realistic train set reconstructions of the classifiers, but rather to produce sample estimations that faithfully represent their manifold. However, we observe that for the sketch modality, all the estimations are quite realistic and akin to what a human amateur sketcher might draw. This is also true for a significant number of reconstructed photos, however for a few (like the Frog), some structural details are missing. This shows that our photo estimators can sometimes be more biased towards replicating textural details over structures, if the former plays a dominant role in the train set of the photo classifier. Also, our estimated photos can be seen to generate distinct objects, but not much background context. We conjecture that this happens because to produce robust predictions, the classifiers themselves learn to be invariant to background information, thereby being unable to provide much feedback about the same.

\subsection{Ablation Studies}

\begin{table}[!t]
    \centering
    \resizebox{\columnwidth}{!}{
    \begin{tabular}{l|c|c|c|c|c|c|c}
        \hline
        \textbf{ID} & \textbf{Semantic} &  \textbf{Class} & \textbf{Modality} & \textbf{Adversarial} & \textbf{Encoding} & \textbf{Sketchy} & \textbf{TU-Berlin} \\
        & \textbf{Consistency} &  \textbf{Alignment} & \textbf{Guidance} & \textbf{Estimation} & \textbf{Loss} & \textbf{(mAP@all)} & \textbf{(mAP@all)} \\
        \hline
        1. & \ding{51} & & & & & 0.600 & 0.416 \\
        2. & \ding{51} & \ding{51} & & & & 0.660 & 0.535 \\
        3. & \ding{51} & \ding{51} & \ding{51} & & & 0.705 & 0.577 \\
        4. & \ding{51} & \ding{51} & \ding{51} & \ding{51} & & 0.740 & 0.630 \\
        \hline
        5. &  & \ding{51} & \ding{51} & \ding{51} & \ding{51} & 0.551 & 0.360 \\
        6. & \ding{51} &  & \ding{51} & \ding{51} & \ding{51} & 0.690 & 0.526 \\
        7. & \ding{51} & \ding{51} &  & \ding{51} & \ding{51} & 0.773 & 0.665\\
        8. & \ding{51} & \ding{51} & \ding{51} & & \ding{51} & 0.758 & 0.641 \\
        \hline
        9. & \ding{51} & \ding{51} & \ding{51} & \ding{51} & \ding{51} & \textbf{0.827} & \textbf{0.680} \\
        \hline
    \end{tabular}}
    \caption{Ablation Studies.}
    \label{tab:ablation}
\end{table}

We present the results of ablating the key components of our model in \cref{tab:ablation}. We group them into two categories. Rows 1 to 4 show the effect of incremental addition of the three novel, domain-specific components of our methodology (from \cref{sec:estimating_distribution}), namely
Class-Alignment, Modality Guidance, and Metric-Agnostic Adversarial Estimation.

Rows 5 to 8 demonstrate how individually removing each of these key components affects the overall performance of the complete model. Row 9 shows the performance of our model with all the components included. We present more details on the individual observations below.

\myparagraph{Semantic Consistency} The semantic consistency loss ($\mathcal{L}_\text{sem}$) is the most fundamental component of a data-free SBIR pipeline \cite{Chen2019DAFL, Choi2020dfAdv, Fang2021dfContr}, which ensures that the reconstructed train set distributions represent meaningful classes. Row-1 represents a baseline model trained only using the semantic consistency loss. The relative contributions of each of our novel components can be evaluated on top of this baseline. Row-5 shows the performance of our model if the semantic consistency criterion is removed. The results show a significant drop in accuracy, as the reconstructed inputs no longer have any meaningful class information and is equivalent to random noise with some modality specific information (coming from the modality guidance network).

\myparagraph{Class-Alignment} The class-alignment loss ($\mathcal{L}_\text{align}$) ensures that the corresponding photos and sketches in the estimated distributions belong to the same class.
Without this, the degree/distribution of class information between a photo-sketch pair may diverge arbitrarily, even if their hard-labels (as predicted by the teacher) happen to be the same. $\mathcal{L}_\text{align}$ ensures more fine-grained correspondence between the pairs by minimizing the symmetrized KL-divergence between the classifiers' outputs. Rows 2 and 6 show that this leads to a significant improvement in retrieval accuracy.

\myparagraph{Modality Guidance} The Modality Guidance Network restricts the estimated distributions to belong to specific modalities, \ie, photos and sketches ($\mathcal{L}_\text{modal}$). This is important for training the encoders, as they need to learn to eliminate irrelevant, modality-specific information, and embed their inputs based on semantic content.
The results in Rows 3 and 7 demonstrate the contribution of the modality guidance network to the process of learning cross-modal invariants in the encoding process.

\myparagraph{Adversarial Estimation} Our proposed technique for Metric-Agnostic Adversarial Estimation of train set distributions directs the estimators towards producing samples that are the hardest for the encoder to correctly encode by communicating feedback across representation spaces through $\mathcal{L}_\text{adv}$. As previously observed in the data-free knowledge distillation literature \cite{Micaelli2019ZeroShotKT, Choi2020dfAdv}, training the encoder on such samples ensures its robustness to semantic variations. Rows 4 and 8 illustrate the significant improvement in the encoders' ability to deal with semantic variations when trained with such samples.

\myparagraph{Encoding Loss}
Based on the observations in the Row groups 1-4 and 5-8, as well as the difference between Rows 4 and 9, we found that the queue-based version of the Info-NCE loss ($\mathcal{L}_\text{enc}$) 
works best for training the encoders with the estimated distributions.
The Triplet loss as an ablative alternative to $\mathcal{L}_\text{enc}$ resulted in degraded performance.

\begin{figure}
    \centering
    \includegraphics[width=0.4\textwidth, height=5cm]{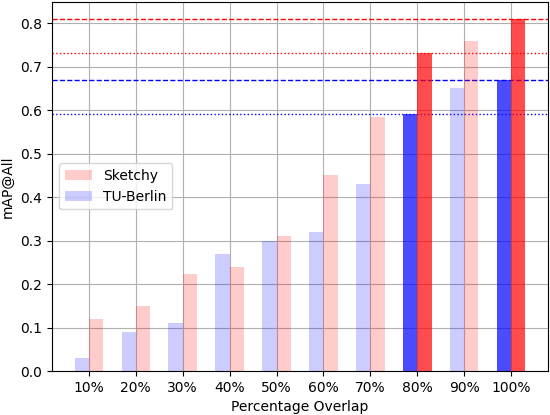}
    \caption{DF-SBIR performance of our model when the classifiers (teachers) are trained on only partially overlapping sets of classes.}
    \label{fig:partial_overlap_graph}
\end{figure}

\subsection{Further Constraints on Classifiers}

\myparagraph{Teachers with Partial Class Overlap} We propose a highly challenging novel setting for the evaluation of cross-modal DFL models. Here, the classifiers (teachers) from the two modalities (photo and sketch) are being trained on \emph{partially overlapping} sets of classes. In other words, not all classes between the photo and the sketch classifiers are common. However, the retrieval has to be performed on the \emph{union} of the training classes of both the classifiers. Formally, if the photo classifier has been trained on the set of classes $\mathcal{C}_p$ and the sketch classifier on $\mathcal{C}_s$, where $\mathcal{C}_p$ and $\mathcal{C}_s$ are partially overlapping, the retrieval test set will constitute of instances from classes $\mathcal{C}_p \bigcup \mathcal{C}_s$.

We make minor modifications to the training process of our model for evaluation under this scenario, the details of which are provided in the supplementary. \cref{fig:partial_overlap_graph} shows its performance as we vary the degree of class overlap between the two classifiers. Even when the encoders of a particular modality do not get direct information about 20\% of the classes (80\% overlap setting), by virtue of our novel class-aligned estimation procedure, our model is able to leverage complementary information across modalities to achieve strong performance in this challenging scenario.

\myparagraph{Cross-Dataset Teachers} To test the ability of our model to deal with the \emph{independence} in the classifiers' training, we evaluate it using photo and sketch classifiers that are each trained on different datasets, but with overlapping classes. With the photo classifier from TU-Berlin and sketch classifier from Sketchy, our model provides an mAP@all of 0.607 on TU-Berlin. With the photo-classifier from Sketchy and sketch classifier from TU-Berlin, we get an mAP@all of 0.759 on Sketchy.
This shows that our model is truly agnostic to the independence in the classifiers' training while learning a cross-modal metric space in a data-free manner.

%% file: tex/05_limitations.tex
\section{Limitations}
\cref{fig:qualitative_results}
shows that our estimators are unable to incorporate much background information in the reconstructions due to lack of feedback from the classifiers, as the classifiers themselves had learned to become invariant to background information for robust prediction performance. However, this is necessary as the downstream encoders also need to
develop this invariance in order to tackle real-world scenarios. We address this by instantiating the encoders with the weights of the classifiers. However, as learning progresses, this knowledge may not be retained given the distribution shift in the estimations. We believe that this limitation can be overcome by leveraging the recent advances in generalized source-free domain adaptation \cite{Yang2021GSFDA}.

%% file: tex/06_concl.tex
\section{Conclusion}

Motivated by the practical constraints of privacy and anonymity preservation, as well as the difficulty involved in obtaining pairs of photos and hand-drawn sketches, for the first time, we studied the problem of SBIR in the data-free setting. We designed an 
approach for performing data-free learning across modalities and metric spaces (CrossX-DFL),
that allows distilling knowledge from independently trained photo and sketch classifiers, into encoders that learn a unified photo-sketch metric space in a completely data-free manner. We illustrate the efficacy of our model
by comparison against an extensive set of 
baselines, as well as exhaustive ablation studies and imposing constraints that challenge its ability
to learn complementary cross-modal information. We hope that our practically motivated problem formulation, along with the methodological design choices that come with it, will propel further research on both DF-SBIR as well as general cross-modal data-free learning.

%% file: main.bbl
\begin{thebibliography}{10}\itemsep=-1pt

\bibitem{Alaniz22SketchPrimitives}
Stephan Alaniz, Massimiliano Mancini, Anjan Dutta, Diego Marcos, and Zeynep
  Akata.
\newblock Abstracting sketches through simple primitives.
\newblock In {\em ECCV}, 2022.

\bibitem{ben_tal2009RobustOptimization}
Aharon Ben-Tal, Laurent Ghaoui, and Arkadi Nemirovski.
\newblock {\em Robust Optimization}.
\newblock Princeton University Press, 2009.

\bibitem{Bhunia_2021_CVPR_MPaA}
Ayan~Kumar Bhunia, Pinaki~Nath Chowdhury, Aneeshan Sain, Yongxin Yang, Tao
  Xiang, and Yi-Zhe Song.
\newblock More photos are all you need: Semi-supervised learning for
  fine-grained sketch based image retrieval.
\newblock In {\em CVPR}, 2021.

\bibitem{bhunia2021sketch2vec}
Ayan~Kumar Bhunia, Pinaki~Nath Chowdhury, Yongxin Yang, Timothy Hospedales, Tao
  Xiang, and Yi-Zhe Song.
\newblock Vectorization and rasterization: Self-supervised learning for sketch
  and handwriting.
\newblock In {\em CVPR}, 2021.

\bibitem{Bhunia2022FewShotInc}
Ayan~Kumar Bhunia, Viswanatha~Reddy Gajjala, Subhadeep Koley, Rohit Kundu,
  Aneeshan Sain, Tao Xiang, and Yi-Zhe Song.
\newblock Doodle it yourself: Class incremental learning by drawing a few
  sketches.
\newblock In {\em CVPR}, 2022.

\bibitem{Bhunia2022NoiseTolerantSBIR}
Ayan~Kumar Bhunia, Subhadeep Koley, Abdullah Faiz Ur~Rahman Khilji, Aneeshan
  Sain, Pinaki~Nath Chowdhury, Tao Xiang, and Yi-Zhe Song.
\newblock Sketching without worrying: Noise-tolerant sketch-based image
  retrieval.
\newblock In {\em CVPR}, 2022.

\bibitem{Chaudhuri2022XModalViT}
Abhra Chaudhuri, Massimiliano Mancini, Yanbei Chen, Zeynep Akata, and Anjan
  Dutta.
\newblock Cross-modal fusion distillation for fine-grained sketch-based image
  retrieval.
\newblock In {\em BMVC}, 2022.

\bibitem{Chen2019DAFL}
Hanting Chen, Yunhe Wang, Chang Xu, Zhaohui Yang, Chuanjian Liu, Boxin Shi,
  Chunjing Xu, Chao Xu, and Qi Tian.
\newblock Dafl: Data-free learning of student networks.
\newblock In {\em ICCV}, 2019.

\bibitem{Cheng2019ImprovingBA}
Shuyu Cheng, Yinpeng Dong, Tianyu Pang, Hang Su, and Jun Zhu.
\newblock Improving black-box adversarial attacks with a transfer-based prior.
\newblock In {\em NeurIPS}, 2019.

\bibitem{Choi2020dfAdv}
Yoojin Choi, Jihwan Choi, Mostafa El-Khamy, and Jungwon Lee.
\newblock Data-free network quantization with adversarial knowledge
  distillation.
\newblock In {\em CVPRW}, 2020.

\bibitem{Chopra2005SiameseLoss}
S. Chopra, R. Hadsell, and Y. LeCun.
\newblock Learning a similarity metric discriminatively, with application to
  face verification.
\newblock In {\em CVPR}, 2005.

\bibitem{Deng2009ImageNetAL}
Jia Deng, Wei Dong, Richard Socher, Li-Jia Li, K. Li, and Li Fei-Fei.
\newblock Imagenet: A large-scale hierarchical image database.
\newblock {\em CVPR}, 2009.

\bibitem{Dey2019CVPR}
Sounak Dey, Pau Riba, Anjan Dutta, Josep Llados, and Yi-Zhe Song.
\newblock Doodle to search: Practical zero-shot sketch-based image retrieval.
\newblock In {\em CVPR}, 2019.

\bibitem{Dutta2019CVPR}
Anjan Dutta and Zeynep Akata.
\newblock Semantically tied paired cycle consistency for zero-shot sketch-based
  image retrieval.
\newblock In {\em CVPR}, 2019.

\bibitem{Dutta2020SEMPCYCAny}
Anjan Dutta and Zeynep Akata.
\newblock Semantically tied paired cycle consistency for any-shot sketch-based
  image retrieval.
\newblock {\em IJCV}, 2020.

\bibitem{eitz2012TUBerlin}
Mathias Eitz, James Hays, and Marc Alexa.
\newblock How do humans sketch objects?
\newblock {\em SIGGRAPH}, 2012.

\bibitem{Fang2021dfContr}
Gongfan Fang, Jie Song, Xinchao Wang, Chengchao Shen, Xingen Wang, and Mingli
  Song.
\newblock Contrastive model inversion for data-free knowledge distillation.
\newblock In {\em IJCAI}, 2021.

\bibitem{federici2020}
Marco Federici, Anjan Dutta, Patrick Forré, Nate Kushman, and Zeynep Akata.
\newblock Learning robust representations via multi-view information
  bottleneck.
\newblock In {\em ICLR}, 2020.

\bibitem{He2020MomentumCF}
Kaiming He, Haoqi Fan, Yuxin Wu, Saining Xie, and Ross~B. Girshick.
\newblock Momentum contrast for unsupervised visual representation learning.
\newblock In {\em CVPR}, 2020.

\bibitem{He2016DeepRL}
Kaiming He, X. Zhang, Shaoqing Ren, and Jian Sun.
\newblock Deep residual learning for image recognition.
\newblock {\em CVPR}, 2016.

\bibitem{hinton2015distilling}
Geoffrey Hinton, Oriol Vinyals, and Jeff Dean.
\newblock Distilling the knowledge in a neural network.
\newblock In {\em NIPSW}, 2015.

\bibitem{hu2022feature}
Conghui Hu and Gim~Hee Lee.
\newblock Feature representation learning for unsupervised cross-domain image
  retrieval.
\newblock In {\em ECCV}, 2022.

\bibitem{Hu2013GFHOG}
Rui Hu and John Collomosse.
\newblock A performance evaluation of gradient field hog descriptor for sketch
  based image retrieval.
\newblock {\em CVIU}, 2013.

\bibitem{Jongejan2016QuickDraw}
Jonas Jongejan, Henry Rowley, Takashi Kawashima, Jongmin Kim, and Nick
  Fox-Gieg.
\newblock The quick, draw! - a.i. experiment.
\newblock \url{https://quickdraw.withgoogle.com}, 2016.

\bibitem{Kim2020CVPR}
Sungyeon Kim, Dongwon Kim, Minsu Cho, and Suha Kwak.
\newblock Proxy anchor loss for deep metric learning.
\newblock In {\em CVPR}, 2020.

\bibitem{Yelamarthi2018ECCV}
Sasi Kiran~Yelamarthi, Shiva Krishna~Reddy, Ashish Mishra, and Anurag Mittal.
\newblock A zero-shot framework for sketch based image retrieval.
\newblock In {\em ECCV}, 2018.

\bibitem{DeformableParts}
Yi Li, Timothy Hospedales, Yi-Zhe Song, and Shaogang Gong.
\newblock Fine-grained sketch-based image retrieval by matching deformable part
  models.
\newblock In {\em BMVC}, 2014.

\bibitem{Liu2017DSH}
L. Liu, F. Shen, Y. Shen, X. Liu, and L. Shao.
\newblock Deep sketch hashing: Fast free-hand sketch-based image retrieval.
\newblock In {\em CVPR}, 2017.

\bibitem{Liu2022CNNFor2020s}
Zhuang Liu, Hanzi Mao, Chao-Yuan Wu, Christoph Feichtenhofer, Trevor Darrell,
  and Saining Xie.
\newblock A convnet for the 2020s.
\newblock In {\em CVPR}, 2022.

\bibitem{lopes2017DFKD}
Raphael~Gontijo Lopes, Stefano Fenu, and Thad Starner.
\newblock Data-free knowledge distillation for deep neural networks.
\newblock {\em arXiv}, 2017.

\bibitem{Lu2022April}
Jiahao Lu, Xi~Sheryl Zhang, Tianli Zhao, Xiangyu He, and Jian Cheng.
\newblock April: Finding the achilles' heel on privacy for vision transformers.
\newblock In {\em CVPR}, 2022.

\bibitem{Lu_2022dfGeomReID}
Yichen Lu, Mei Wang, and Weihong Deng.
\newblock Augmented geometric distillation for data-free incremental person
  reid.
\newblock In {\em CVPR}, 2022.

\bibitem{Mahendran2015DFInterpret}
Aravindh Mahendran and Andrea Vedaldi.
\newblock Understanding deep image representations by inverting them.
\newblock In {\em CVPR}, 2015.

\bibitem{McMahan2017CommunicationEfficientLO}
H.~B. McMahan, Eider Moore, Daniel Ramage, Seth Hampson, and Blaise~Ag{\"u}era
  y Arcas.
\newblock Communication-efficient learning of deep networks from decentralized
  data.
\newblock In {\em AISTATS}, 2017.

\bibitem{Micaelli2019ZeroShotKT}
Paul Micaelli and Amos~J Storkey.
\newblock Zero-shot knowledge transfer via adversarial belief matching.
\newblock In {\em NeurIPS}, 2019.

\bibitem{Movshovitz-Attias2017ProxyNCA}
Yair Movshovitz-Attias, Alexander Toshev, Thomas~K. Leung, Sergey Ioffe, and
  Saurabh Singh.
\newblock No fuss distance metric learning using proxies.
\newblock In {\em ICCV}, 2017.

\bibitem{BMVC2017_46}
Kaiyue Pang, Yi-zhe Song, Tony Xiang, and Timothy Hospedales.
\newblock Cross-domain generative learning for fine-grained sketch-based image
  retrieval.
\newblock In {\em BMVC}, 2017.

\bibitem{PangKaiyue2020SMJP}
Kaiyue Pang, Yongxin Yang, Timothy~M Hospedales, Tao Xiang, and Yi-Zhe Song.
\newblock Solving mixed-modal jigsaw puzzle for fine-grained sketch-based image
  retrieval.
\newblock In {\em CVPR}, 2020.

\bibitem{Park2019RelationalKD}
Wonpyo Park, Dongju Kim, Yan Lu, and Minsu Cho.
\newblock Relational knowledge distillation.
\newblock In {\em CVPR}, 2019.

\bibitem{PretrainedModelsPyTorch}
Pytorch.
\newblock Pre-trained models in pytorch.
\newblock \url{https://pytorch.org/vision/stable/models.html}.

\bibitem{Roth2021S2SD}
Karsten Roth, Timo Milbich, Bjorn Ommer, Joseph~Paul Cohen, and Marzyeh
  Ghassemi.
\newblock Simultaneous similarity-based self-distillation for deep metric
  learning.
\newblock In {\em ICML}, 2021.

\bibitem{Saavedra2014SHELO}
Jose~M Saavedra.
\newblock Sketch based image retrieval using a soft computation of the
  histogram of edge local orientations (s-helo).
\newblock In {\em ICIP}, 2014.

\bibitem{Saavedra2015LKS}
Jose~M Saavedra, Juan~Manuel Barrios, and S Orand.
\newblock {Sketch based Image Retrieval using Learned KeyShapes (LKS)}.
\newblock In {\em BMVC}, 2015.

\bibitem{Sain_2021_CVPR_StyleMeUp}
Aneeshan Sain, Ayan~Kumar Bhunia, Yongxin Yang, Tao Xiang, and Yi-Zhe Song.
\newblock Stylemeup: Towards style-agnostic sketch-based image retrieval.
\newblock In {\em CVPR}, 2021.

\bibitem{Sangkloy2016Sketchy}
Patsorn Sangkloy, Nathan Burnell, Cusuh Ham, and James Hays.
\newblock The sketchy database: Learning to retrieve badly drawn bunnies.
\newblock {\em ACM SIGGRAPH}, 2016.

\bibitem{Sanyal2022dfModelStealing}
Sunandini Sanyal, Sravanti Addepalli, and R.~Venkatesh Babu.
\newblock Towards data-free model stealing in a hard label setting.
\newblock In {\em CVPR}, 2022.

\bibitem{FgsbirSpatialAttention}
Jifei Song, Qian Yu, Yi-Zhe Song, Tao Xiang, and Timothy~M. Hospedales.
\newblock Deep spatial-semantic attention for fine-grained sketch-based image
  retrieval.
\newblock In {\em ICCV}, 2017.

\bibitem{Tian2020Contrastive}
Yonglong Tian, Dilip Krishnan, and Phillip Isola.
\newblock Contrastive representation distillation.
\newblock In {\em ICLR}, 2020.

\bibitem{Oord2018RepresentationLW}
A{\"a}ron van~den Oord, Yazhe Li, and Oriol Vinyals.
\newblock Representation learning with contrastive predictive coding.
\newblock {\em arXiv}, 2018.

\bibitem{wang2019ImageNetSketch}
Haohan Wang, Songwei Ge, Zachary Lipton, and Eric~P Xing.
\newblock Learning robust global representations by penalizing local predictive
  power.
\newblock In {\em NeurIPS}, 2019.

\bibitem{Wang2022DST}
Wenxuan Wang, Xuelin Qian, Yanwei Fu, and Xiangyang Xue.
\newblock Dst: Dynamic substitute training for data-free black-box attack.
\newblock In {\em CVPR}, 2022.

\bibitem{Wortsman2022ModelSoups}
Mitchell Wortsman, Gabriel Ilharco, Samir~Ya Gadre, Rebecca Roelofs, Raphael
  Gontijo-Lopes, Ari~S Morcos, Hongseok Namkoong, Ali Farhadi, Yair Carmon,
  Simon Kornblith, and Ludwig Schmidt.
\newblock Model soups: averaging weights of multiple fine-tuned models improves
  accuracy without increasing inference time.
\newblock In {\em ICML}, 2022.

\bibitem{Wu2018CVPR}
Zhirong Wu, Yuanjun Xiong, Stella~X. Yu, and Dahua Lin.
\newblock Unsupervised feature learning via non-parametric instance
  discrimination.
\newblock In {\em CVPR}, 2018.

\bibitem{Xie_2020_CVPR}
Qizhe Xie, Minh-Thang Luong, Eduard Hovy, and Quoc~V. Le.
\newblock Self-training with noisy student improves imagenet classification.
\newblock In {\em CVPR}, 2020.

\bibitem{Yang2021GSFDA}
Shiqi Yang, Yaxing Wang, Joost van~de Weijer, Luis Herranz, and Shangling Jui.
\newblock Generalized source-free domain adaptation.
\newblock In {\em ICCV}, 2021.

\bibitem{Yin2020DreamingToDistill}
Hongxu Yin, Pavlo Molchanov, Jose~M. Alvarez, Zhizhong Li, Arun Mallya, Derek
  Hoiem, Niraj~K. Jha, and Jan Kautz.
\newblock Dreaming to distill: Data-free knowledge transfer via deepinversion.
\newblock In {\em CVPR}, 2020.

\bibitem{Yoo2019KnowledgeEW}
Jaemin Yoo, Minyong Cho, Taebum Kim, and U Kang.
\newblock Knowledge extraction with no observable data.
\newblock In {\em NeurIPS}, 2019.

\bibitem{Yu2016SketchMe}
Qian Yu, Feng Liu, Yi-Zhe Song, Tao Xiang, Timothy~M. Hospedales, and
  Chen~Change Loy.
\newblock Sketch me that shoe.
\newblock In {\em CVPR}, 2016.

\bibitem{Zhang_2022dfAttack}
Jie Zhang, Bo Li, Jianghe Xu, Shuang Wu, Shouhong Ding, Lei Zhang, and Chao Wu.
\newblock Towards efficient data free black-box adversarial attack.
\newblock In {\em CVPR}, 2022.

\bibitem{Zhang2022dfFedLearn}
Lin Zhang, Li Shen, Liang Ding, Dacheng Tao, and Ling-Yu Duan.
\newblock Fine-tuning global model via data-free knowledge distillation for
  non-iid federated learning.
\newblock In {\em CVPR}, 2022.

\bibitem{Zhang2021DFSuperRes}
Yiman Zhang, Hanting Chen, Xinghao Chen, Yiping Deng, Chunjing Xu, and Yunhe
  Wang.
\newblock Data-free knowledge distillation for image super-resolution.
\newblock In {\em CVPR}, 2021.

\end{thebibliography}


\begin{thebibliography}{1}\itemsep=-1pt

\bibitem{Deng2009ImageNetAL}
Jia Deng, Wei Dong, Richard Socher, Li-Jia Li, K. Li, and Li Fei-Fei.
\newblock Imagenet: A large-scale hierarchical image database.
\newblock {\em CVPR}, 2009.

\bibitem{federici2020}
Marco Federici, Anjan Dutta, Patrick Forré, Nate Kushman, and Zeynep Akata.
\newblock Learning robust representations via multi-view information
  bottleneck.
\newblock In {\em ICLR}, 2020.

\bibitem{He2016DeepRL}
Kaiming He, X. Zhang, Shaoqing Ren, and Jian Sun.
\newblock Deep residual learning for image recognition.
\newblock {\em CVPR}, 2016.

\bibitem{Karras2020StyleGAN2}
Tero Karras, Samuli Laine, Miika Aittala, Janne Hellsten, Jaakko Lehtinen, and
  Timo Aila.
\newblock Analyzing and improving the image quality of stylegan.
\newblock In {\em CVPR}, 2020.

\bibitem{Liu2017DSH}
L. Liu, F. Shen, Y. Shen, X. Liu, and L. Shao.
\newblock Deep sketch hashing: Fast free-hand sketch-based image retrieval.
\newblock In {\em CVPR}, 2017.

\bibitem{lopes2017DFKD}
Raphael~Gontijo Lopes, Stefano Fenu, and Thad Starner.
\newblock Data-free knowledge distillation for deep neural networks.
\newblock {\em arXiv}, 2017.

\bibitem{Paszke2017PyTorch}
Adam Paszke, Sam Gross, Soumith Chintala, Gregory Chanan, Edward Yang, Zachary
  DeVito, Zeming Lin, Alban Desmaison, Luca Antiga, and Adam Lerer.
\newblock Automatic differentiation in {PyTorch}.
\newblock In {\em NIPSW}, 2017.

\end{thebibliography}
